# Estimation of Variance and Spatial Correlation Width for Fine-scale Measurement Error in Digital Elevation Model

M.L. Uss, B. Vozel, V.V. Lukin, K. Chehdi

*Abstract*— In this paper, we borrow from blind noise parameter estimation (BNPE) methodology early developed in the image processing field an original and innovative no-reference approach to estimate Digital Elevation Model (DEM) vertical error parameters without resorting to a reference DEM. The challenges associated with the proposed approach related to the physical nature of the error and its multifactor structure in DEM are discussed in detail. A suitable multivariate method is then developed for estimating the error in gridded DEM. It is built on a recently proposed vectorial BNPE method for estimating spatially correlated noise using Noise Informative areas and Fractal Brownian Motion. The newly multivariate method is derived to estimate the effect of the stacking procedure and that of the epipolar line error on local (fine-scale) standard deviation and autocorrelation function width of photogrammetric DEM measurement error. Applying the new estimator to ASTER GDEM2 and ALOS World 3D DEMs, good agreement of derived estimates with results available in the literature is evidenced. In future works, the proposed no-reference method for analyzing DEM error can be extended to a larger number of predictors for accounting for other factors influencing remote sensing (RS) DEM accuracy.

## I. INTRODUCTION

GRIDDED DEMs have found applications for disaster and crisis-management support [1], urban growth monitoring and planning [2], RS image processing [3]. A DEM is a subject to error that may impair its quality. This error originates from various sources, including elevation measurement method (traditional optical stereo matching, radar interferometry (IfSAR), or Light Detection and Ranging (LiDAR) [4]), instrument, terrain structure, vegetation cover, and DEM interpolation, making both theoretical and experimental analysis of its properties a complicated task [5]. Only vertical DEM error is of interest in this paper. In what follows, the multifactor nature of DEM error is referred to as multivariate context-dependence.

DEM error leads to uncertainty in calculation of terrain attributes such as terrain slope, aspect, or roughness. Even partial knowledge of DEM error properties and its spatial pattern is valuable [5-7] in such areas as DEM fusion [8, 9], filling voids [10], DEM filtering [7, 11-13], modeling of DEM error propagation [14] in hillslope erosion/failure analysis, land slide risk estimation, hydrological modeling. Improving knowledge of DEM error was identified as a major research direction in digital terrain modeling domain [15].

The basic DEM error characteristic [7] is its 
$$\text{RMSE} = \sqrt{\frac{1}{n}\sum_{i=1}^{n}(z_{\text{DEM}.i} - z_{\text{Ref}.i})^2}, \text{ where } z_{\text{DEM}.i} \text{ is an}$$
elevation measurement from the DEM, $z_{\text{Ref}.i}$ is the reference elevation measurement of significantly higher accuracy, $n$ is number of available measurements. RMSE can be decomposed as $RMSE = \sqrt{\sigma_e^2 + M_e^2}$, where mean error or bias $M_e = \frac{1}{n}\sum_{i=1}^{n}(z_{\text{DEM}.i} - z_{\text{Ref}.i})$ and standard deviation (SD) $\sigma_e = \sqrt{\frac{1}{n}\sum_{i=1}^{n}(z_{\text{DEM}.i} - z_{\text{Ref}.i} - M_e)^2}$ [5]. The $\sigma_e$ and $M_e$ terms characterize random and systematic error components, respectively. Spatial properties of DEM error are characterized by its spatial correlation function [14].

To measure DEM error characteristics, a high quality reference DEM (e.g. created using LIDAR) or point elevation measurements (obtained by geodetic surveys) are typically utilized [6, 16, 17]. Comparison between DEMs is a complicated task. Both analyzed and reference DEMs should have the same spatial resolution, cover the same area and share approximately the same acquisition date. The reference DEM should have significantly better accuracy than the analyzed DEM. Acquisition of such data can be expensive and is not always possible. Direct DEM comparison is further complicated for vegetation cover that can yield different elevation measurements for different instruments (optical,

M. L. Uss is with Department of Transmitters, Receivers and Signal Processing, National Aerospace University, Kharkov, Ukraine (e-mail: uss@xai.edu.ua).

B. Vozel is with IETR UMR CNRS 6164 - University of Rennes 1, Enssat Lannion, France (phone: 33 296469071; fax: 33 296469075; e-mail: benoit.vozel@univ-rennes1.fr).

V. V. Lukin is with the Department of Receivers, Transmitters and Signal Processing, National Aerospace University, Kharkov, Ukraine (email: lukin@ai.kharkov.com).

K. Chehdi is with IETR UMR CNRS 6164 - University of Rennes 1, Enssat Lannion, France (e-mail: kacem.chehdi@univ-rennes1.fr).

InSAR, LIDAR) [18-20].

Above-mentioned tight requirements on the reference DEM can be relaxed for featureless flat terrain. For a local DEM patch representing a flat terrain, $z_{\text{Ref}} = const$ and DEM SD can be directly accessed as $\sigma_e = \sqrt{\frac{1}{n}\sum_{i=1}^{n}\left(z_{\text{DEM},i} - \frac{1}{n}\sum_{i=1}^{n}z_{\text{DEM},i}\right)^2}$. In this case, reference DEM is not needed to estimate $\sigma_e$ (however, it is still needed to estimate DEM bias). Having enough flat patches allows collecting many local $\sigma_e$ estimates for DEM accuracy characterization. Importantly, using flat DEM patches, SD of DEM error can be estimated in no-reference fashion. This approach was leveraged by K. Becek in [20]. He proposed to use runways as flat patches and estimated bias and SD of the ASTER Global DEM2 [21, 22] (later referred to as GDEM2) and Shuttle Radar Topography Mission (SRTM) DEM [23] error.

Usage of flat terrain encounters several problems. First, flat areas are unknown beforehand (with narrow exception exploited by K. Becek). Second, absolutely flat areas are hard to find. Let us consider estimation of $\sigma_e$ value locally from a DEM patch of 15 by 15 pixels. For GDEM2, one pixel corresponds to 30m on the ground; the patch is thus 450 by 450m. The mean $\sigma_e$ value for GDEM2 has been shown in [20] to vary from 2 to 6m depending on stacking number. To reliably estimate a low value of $\sigma_e$, such as $\sigma_e = 2\text{m}$, the standard deviation of underlying terrain elevation should be a magnitude smaller, e.g. 0.2m. This implies maximum elevation variation ±0.6m within the patch. This restriction severely reduces number and variability of flat patches available for no-reference DEM error characterization.

The process of estimation of DEM error SD from featureless flat terrain has a direct counterpart in image processing domain, namely blind noise parameters estimation approach [24], where the goal is to estimate image sensor noise characteristics (variance or autocorrelation function) from a mixture of true signal and noise. The common solution is to find and use image Homogeneous Areas (HA), where true signal is negligible as compared to sensor noise [25]. The link between no-reference DEM error characterization and BNPE problem is that flat terrain for DEM corresponds to HA for images, DEM error - to sensor noise, and true signal to error-free DEM. The problem of automatic HA search is the core of BNPE approach, and many efficient solutions have been proposed. Furthermore, BNPE methodology moved beyond searching HA areas only: moderately heterogeneous areas could be used for noise parameters estimation by assuming spectral differences between noise and the true signal [26]. Therefore, BNPE methodology provides suitable approaches to two main barriers to no-reference DEM error characterization discussed above: automatic search of HA (flat DEM patches) and usage of moderately heterogeneous patches (DEM patches corresponding to moderately undulating terrain).

Going through obvious similarities underlined above, adapting key-ideas from BNPE methodology so as to make them fit the issues and problems to solve for gridded DEM is the main contribution of this paper. The main difficulty lies in multivariate context-dependence of DEM error compared to univariate dependence of sensor noise on image intensity widely considered in BNPE domain (this issue is discussed more in detail in the next section). To the best of our knowledge, the only blind noise parameter estimator designed to deal with multivariate context-dependent error was proposed by authors of this paper in [26] and called mvcNI+fBm (Multivariate, Vector estimator of spatially Correlated noise using Noise Informative areas and Fractal Brownian Motion). Therefore, in this paper, we modify and evolve mvcNI+fBm estimator to make it adjusted to DEM data and apply it to two global photogrammetric gridded DEMs: GDEM2 and ALOS Global Digital Surface Model "ALOS World 3D" with 5 m spacing (AW3D) [27] and 30 m spacing (AW3D30) [28]. In regard to the complexity of DEM error as compared to pure sensor noise, we devote significant efforts to explain and interpret the meaning of estimates obtained with the proposed approach.

The remainder of this paper is structured as follows. In Section II, we start by discussing possible sources of error in DEM. Then, we establish that sensitivity to fine-scale error offered by BNPE methodology is fully beneficial to fine-scale elevation measurement error. Section III introduces mathematical model of the DEM measurement error and several bivariate regression models for the measurement error variance and spatial correlation width. Section IV details necessary information on the mvcNI+fBm estimator. In Section V, the experimental section, the mvcNI+fBm is used to select the most relevant measurement error model and estimate its coefficients using GDEM2, AW3D30, and AW3D data. It is demonstrated that obtained DEM error models are physically adequate and in good agreement with the respective DEM accuracy analysis published in the literature. On the basis of these promising results, concluding remarks and future work are given.

## II. DEM ERROR MODEL

An instrument (sensor) measures elevation at discrete, possibly irregular spaced points on the Earth surface. To form a gridded DEM, discrete measurements are interpolated at nodes on a regular grid: $\hat{Z}_{grid}(x_t, y_s) = \hat{Z}_{grid}(t,s)$, where $x_t = x_0 + t \cdot r_{\text{DEM}}$, $y_s = y_0 + s \cdot r_{\text{DEM}}$, $r_{\text{DEM}}$ is the grid step or DEM spatial resolution, $(x_0, y_0)$ is the grid origin.

The measured elevation is a coarsened representation of the actual terrain because of the finite instrument spatial resolution (correlation window support in photogrammetric DEM, sounding beam footprint in InSAR and LIDAR). Elevation measurement process is subject to measurement errors (Fig. 1). For bare soil, the instrument introduces error related to its positioning/orientation accuracy, disparity estimation errors, higher or lower correlation between stereo images, stacking number, epipolar line error in photogrammetric DEM, phase unwrapping errors in InSAR, time-of-delay measurement accuracy in LIDAR [5, 6, 20, 29, 30]. This error source was called instrument-induced in [20].

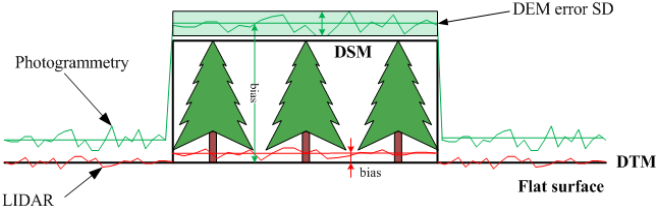

Fig. 1. Illustration of DEM error structure.

The bias and SD of this error could depend on terrain morphological attributes such as slope, aspect, roughness [7]. Vegetation canopy is additional source of DEM error [29]. For example, optical sensors measure elevation at the canopy top, LIDAR could measure both the vegetation canopy top (the first return) and ground surface (the last return), InSAR measures intermediate elevation between the vegetation canopy top and the ground surface [7, 17]. In [4], it was demonstrated that both bias and SD of SRTM DEM error significantly increase in areas covered by natural forests as compared to grassland pastures and agricultural areas. The error related to terrain surface properties was called environment-induced in [20]. Finally, interpolation and quantization errors are added. Gridded DEM error may exhibit spatial correlation that should be taken into account as well [5, 6, 30]. In what follows, we refer to factors/attributes influencing DEM error as error predictors and denote them as $p_1, p_2, ..., p_{N_p}$, where $N_p$ is number of predictors. In vector form, predictor vector is defined as $\mathbf{p} = (p_1, p_2, ..., p_{N_p})$. Dependence of DEM error parameters (SD in particular) on multiple predictors is the essence of multivariate context-dependence.

Therefore, values of gridded DEM can be represented as $\hat{Z}_{grid}(t,s) = Z_{grid}(t,s) + e_{grid.mes}(t,s,\mathbf{p}(t,s))$, where $Z_{grid}(t,s)$ is error-free DEM and $e_{grid.mes}(t,s,\mathbf{p}(t,s))$ is DEM measurement error. In this paper, we treat the difference $Z_{true}(t,s) - Z_{grid}(t,s)$ between the actual elevation $Z_{true}(t,s)$ and error-free DEM not as error, but as uncertainty [5] dependent on the relation between DEM spatial resolution and terrain characteristic size and DEM interpolation method. By DEM error analysis we understand the study of $e_{grid.mes}(t,s,\mathbf{p}(t,s))$ measurement error term.

To evaluate DEM error, many researchers have used residuals $r(t,s) = \hat{Z}_{grid}(t,s) - \hat{Z}_{grid.ref}(t,s)$ between the analyzed DEM and a more accurate reference DEM. The drawback of this approach is that reference DEM has its own errors that not always could be neglected [31]. The residual $r(t,s) = [Z_{grid}(t,s) - Z_{grid.ref}(t,s)] + [e_{grid.mes}(t,s,\mathbf{p}(t,s)) - e_{grid.ref.mes}(t,s,\mathbf{p}_{ref}(t,s))]$ comprises two error terms. For the first term – difference between error-free DEMs – to be negligible, both DEMs should have the same spatial resolution and perfect spatial alignment (absence of planimetric errors) [32]. The second term includes errors of both analyzed and reference DEMs. To minimize influence of the reference DEM error, this DEM should be obtained with an order of magnitude accurate instrument free from vegetation canopy and terrain surface parameters influence. Analysis of residuals complicates further if analyzed and reference DEM were collected with a time lag because of terrain dynamics, vegetation change (leaf-on, leaf-off conditions [31]) among other factors. The problem remains how to characterize error of an accurate DEM when finding a sufficiently accurate reference DEM is impossible.

In this paper, we investigate the possibility of characterizing DEM error without the use of any reference DEM, but relying on noisy measurements $\hat{Z}_{grid}$ themselves. This problem is known in image processing domain as blind noise parameter estimation problem and it basically aims at estimating random noise characteristics (variance, autocorrelation function parameters) from a mixture of noise-free image and noise, i.e. noisy image [24]. From image processing point of view, the DEM $\hat{Z}_{grid}$ is a noisy single-component image, $Z_{grid}$ is the noise-free image, and $e_{grid.mes}$ is the noise term. In what follows, we analyze what terms of DEM error are accessible by BNPE. Such a preliminary analysis is essential to correctly interpret the results obtained by the mvcNI+fBm estimator derived for real DEMs.

To illustrate better the connection between DEM error characterization and BNPE problems, let us discuss more in detail the "runway" method proposed by K. Bacek in [20]. According to it, if we assume that for a particular DEM patch the reference DEM is flat, the variance of residual between analyzed and reference DEMs is composed of instrumental ($\sigma_I^2$), environmental ($\sigma_E^2$) components and undesirable target-induced component $\sigma_T^2$ that is caused by the unaccounted terrain roughness: $\sigma_Z^2 = \sigma_I^2 + \sigma_E^2 + \sigma_T^2$. The value $\sigma_T^2$ can be approximated as $\sigma_T^2 = \frac{1}{12} r_{DEM}^2 \tan^2(\alpha) + \frac{1}{12} q^2$, where $\alpha$ is terrain root mean square slope, $q$ is quantization step [20]. For a flat terrain (e.g., runway), $\alpha \approx 0$ and target-induced term becomes negligible. In the above introduced terms, for flat terrain $Z_{grid}(t,s) = Z_{grid} = const$, $\hat{Z}_{grid.ref}(t,s) = Z_{grid.ref}(t,s) = Z_{grid.ref} = const$, $e_{grid.ref.mes}(t,s,\mathbf{p}_{ref}(t,s)) = 0$, and $r(t,s)$ simplifies to $[Z_{grid} - Z_{grid.ref}] + e_{grid.mes}(t,s,\mathbf{p}(t,s))$. Even not knowing the actual evaluation $Z_{grid.ref}$, variance of DEM error can be directly estimated as

$$Var(e_{grid.mes}(t,s,\mathbf{p}(t,s))) = \\ = Var(\hat{Z}_{grid}(t,s) - Z_{grid.ref}) = Var(\hat{Z}_{grid}(t,s)) \quad (1)$$

This idea directly corresponds to the so-called HA approach, which is the simplest BNPE approach [25]. The limiting factors of the "runway" method is that (i) it still relies on reference data to locate flat terrain patches; (ii) number of flat patches provided by this method is limited; (iii) these patches represent only one particular terrain class - concrete surfaces, and cannot, in principle, be extended to other terrain classes; (iiii) expression (1) requires absolutely flat patches that is hard to satisfy. These limitations have been overcome in advanced BNPE methods that provide means for automatic search of HA and ability to deal with moderately heterogeneous patches. Thus, for global DEMs, BNPE could potentially rely on a larger amount of data for analyzing DEM error and cover larger variety of DEM error predictors.

BNPE has seen fast development in the last decade and now provides a mature set of methods applicable to a variety of scenarios [24]: single channel [33], multispectral (or color) [34] and hyperspectral images [35, 36]; optical [37] and radar [38] images; signal-independent (additive) [33], Poisson [39], multiplicative noise [38], or their mixture in the form of general model with signal-dependent noise variance [34, 40-42]; methods with ability to characterize only noise variance [33-41] as well as noise spatial correlation properties [42, 43]. Accuracy of these methods is high enough to consider them as an alternative to direct sensor calibration [34-36, 39, 41].

The BNPE problem is ill-conditioned and requires additional *a priori* information on noise and noise-free image properties to be solved reliably. Such additional information could originate from either spatial [25] or spectral [44] domains leading to two main groups of blind noise parameter estimators working in spatial or spectral domain, respectively. Spatial *a priori* information assumes that noise and noise-free image could be separated in image HA where noise-free image level variation is negligible as compared to noise. Spectral *a priori* information assumes that noise-free image and noise difference in spectral domain could help their separation. State-of-the-art methods typically utilize both spatial and spectral information; the mvcNI+fBm method also belongs to this group. Irrespectively from *a priori* information used, blind noise parameter estimators operate by finding image patches where noise-free image and noise can be separated in the best way and locally estimating noise variance from such patches [24].

To illustrate how moderately heterogeneous patches can be used for BNPE problem and which components of DEM error can be estimated by this approach it is useful to represent BNPE as a high-pass filter. The value of

$$Var\left(\hat{Z}_{grid}(t,s)\right) = \frac{1}{N^2}\sum_{i,j=1}^{i,j\leq N}\left(\hat{Z}_{grid}(t,s) - \frac{1}{N^2}\sum_{i,j=1}^{i,j\leq N}\hat{Z}_{grid}(t,s)\right)^2$$

is local variance of residuals $\hat{Z}_{grid}(t,s) - \frac{1}{N^2}\sum_{i,j=1}^{i,j\leq N}\hat{Z}_{grid}(t,s)$ between DEM $\hat{Z}_{grid}(t,s)$ and its low-pass filtered version $\frac{1}{N^2}\sum_{i,j=1}^{i,j\leq N}\hat{Z}_{grid}(t,s)$. Here $N$ is linear patch size. The difference between signal and its low-pass filtered version is equivalent to a high-passed filter applied to the signal. The error-free elevation has multiscale property and possesses fine-, meso- as well as macroscale variations [45]. In contrast, instrument-induced error has slow-varying bias component caused by instrument calibration errors (orientation error, jitter noise) and exhibits random fluctuations at the fine-scale.

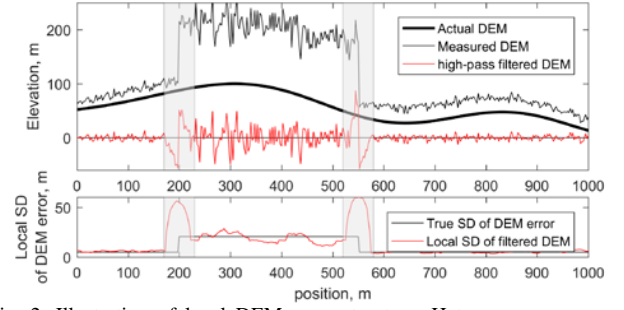

Fig. 2. Illustration of local DEM error structure. Heterogeneous areas are marked by grey fill.

Environment-induced error is related to the terrain surface and could borrow at some extent its multiscale structure. High-pass filter suppresses constant term and large-scale elevation variations and keeps only fine-scale details. This procedure partly removes error-free DEM variations making some of moderately heterogeneous patches almost homogeneous and suitable for estimating DEM error through BNPE; the error-free DEM and error bias terms are also removed as well as low-frequency calibration error; large-scale environment-induced variations of DEM error (bias between bare soil DEM and terrain with vegetation cover) are likewise filtered-out. In contrast, fine-scale random variations of DEM error (both instrument- and environment-induced) pass through high-pass filter. This is illustrated in Fig. 2 where a terrain surface and DEM error are simulated. After high-pass filtering, HA areas represent fine-scale component of DEM error. Heterogeneous areas (shown by grey color in Fig. 2) do not represent DEM error, should be detected and removed from further consideration. In other words, BNPE can be viewed as a filtering process that separates a DEM error from the actual DEM using difference in their autocorrelation functions.

Let us summarize BNPE properties important for understanding effect of their application to gridded DEM: (1) it does not require any reference image (DEM), (2) it characterizes image noise (DEM error) locally, (3) it employs spatial (HAs) and spectral (different autocorrelation function of the true signal and noise) separability of true signal (error-free DEM) and noise (DEM error), (4) it performs automatic search of image patches where the true signal (error-free DEM) and noise (DEM error) are separable at an extent allowing noise (DEM error) parameters estimation, (5) it estimates only fine-scale random noise (DEM error) component, bias and large-scale components cannot be estimated. Therefore, BNPE methods are able to estimate local SD of fine-scale component of DEM measurement error. This error is similar to the residual error or leveled error introduced in [46], that indicates the best possible DEM error with no instrument orientation error affecting the result.

In the majority of BNPE methods, noise can be represented as the sum of two terms: a signal-independent or additive and signal-dependent one. In RS applications, the additive noise is sensor noise (e.g. thermal noise) and it does not depend on the sensed image. The signal-dependent noise is due to the physical nature of measured quantity (e.g. photon-counting or Poisson noise in optical images, and coherent speckle noise in

SAR images) and its variance is a linear or non-linear function of sensed image intensity [47, 48]. Therefore, this observation noise model is univariate. In contrast, DEM error depends on many predictors both instrument- and environment-related. To the best of our knowledge, the only BNPE method able to deal with multivariate noise model is the recently proposed mvcNI+fBm (see Section IV for details). This property has determined our choice in favor of this estimator for characterizing DEM error.

Including all significant predictors selected for analysis in predictor vector $\mathbf{p}$, one expect DEM error to exhibit the same statistical properties for all patches with the same predictor vector. However, not all predictors could be observed, identified or considered in a particular study. Therefore, predictor vector is split into observable and not-observable, latent, parts: $\mathbf{p} = (\mathbf{p}_{observed}, \mathbf{p}_{latent})$. The mvcNI+fBm is designed to estimate measurement error SD $\sigma_e$ for patches with similar observable predictors $\mathbf{p}_{observed}$. However, latent predictors are not controlled leading to variations of the estimates SD: $\hat{\sigma}_{e.i}^2 = \hat{\sigma}_e^2(\mathbf{p}_{observed}, \mathbf{p}_{latent.i})$, where $i = 1...N_{pt}$, $N_{pt}$ is number of patches. In the simplest case, final SD estimates are obtained as average over all patches:

$$\hat{\sigma}_e^2 = N_{pt}^{-1} \sum_{i=1}^{N_{pt}} \hat{\sigma}_{e.i}^2(\mathbf{p}_{observed}, \mathbf{p}_{latent.i}) \approx \hat{\sigma}_e^2(\mathbf{p}_{observed}, \bar{\mathbf{p}}_{latent}),$$

where $\bar{\mathbf{p}}_{latent} = N_{pt}^{-1} \sum_{i=1}^{N_{pt}} \mathbf{p}_{latent.i}$. The mvcNI+fBm estimates error parameters for a fixed value of observable predictors and an average value of the latent ones.

Significant limitation of BNPE resides in its operation principle. As DEM error SD is estimated from homogeneous or moderately heterogeneous DEM patches representing low relief terrain, predictor vector values that correspond to high relief terrain are inaccessible. Primarily, DEM error parameters for patches with high slope SD cannot be estimated using no-reference approach. Homogeneity level required for reliable DEM error SD estimation is related to the DEM error SD itself: the lower DEM error is, the more homogeneous patches are needed to access it. The DEM error component accessible by the BNPE methods can be finally formulated as fine-scale measurement error SD in low-relief areas averaged over latent, unobservable predictors. In what follows, we refer to this DEM error component simply as measurement error or even error, omitting other qualifiers.

### III. ELEVATION MEASUREMENT ERROR MODEL

*A. Mathematical model of DEM error*

Following the discussion in previous section, for an $(2N_h+1) \times (2N_h+1)$ DEM patch, we assume the following model:

$$\hat{Z}_{grid}(t,s) = Z_{grid}(t,s) + e_{Hgrid.mes}(t,s,\mathbf{p}), \quad (2)$$

where $t, s = -N_h...N_h$ denote pixel coordinates within the patch, $N_h$ is the patch half size, $N = 2N_h+1$ is the patch linear size, $\hat{Z}_{grid}(t,s)$ and $Z_{grid}(t,s)$ are affected by errors and error-free DEMs, $e_{grid.mes}(t,s,\mathbf{p})$ is a zero-mean spatially correlated random error, that models the fine-scale measurement error. For a particular patch, predictor vector $\mathbf{p}$ is assumed constant, but it changes its value from patch to patch. Model (2) resembles DEM modeling in geostatistics when a DEM patch is split into zero-mean error term and spatially autocorrelated random field modeling error-free DEM [7].

It is assumed that error is uncorrelated with the error-free DEM. By this we mean that only error parameters are dependent on DEM structure and not error realization itself, i.e., error and error-free elevation values in the same pixel have no linear relationship with each other. The Gaussian distribution is the most natural assumption for DEM error, which is generated by a complex mixture of many factors. The validity of this distribution has been reported in many sources [4-6, 9, 30].

Besides, the semivariogram analysis of DEM error indicates presence of spatial autocorrelation [6, 30]. At the lag distances considered in this paper (below 100-200 m), isotropy of DEM errors is also typically assumed (anisotropy of DEM errors was reported in [30] for lag distances exceeding 500 m). With this, no strong evidence supporting a particular spatial autocorrelation function shape is present in the available literature. We performed experiments with two widely used correlation function shapes: exponential

$$R_e(t_1, s_1, t_2, s_2) = \sigma_e^2(\mathbf{p}) \exp\left(-\frac{|d|}{\sigma_{Corr}(\mathbf{p})}\right) \quad [14], \text{ and Gaussian}$$

$$R_e(t_1, s_1, t_2, s_2) = \sigma_e^2(\mathbf{p}) \exp\left(-\frac{d^2}{2\sigma_{Corr}^2(\mathbf{p})}\right), \quad \text{where}$$

$d = \sqrt{(t_2-s_1)^2 + (t_2-s_1)^2}$, $\sigma_e^2(\mathbf{p})$ is the noise variance, and $\sigma_{Corr}^2(\mathbf{p})$ is the parameter that characterizes unknown error spatial correlation width ($\sigma_{Corr}$ is distance in pixels where error correlation drops to 0.6). It was found that Gaussian shape is more adequate than exponential one (in terms of regression quality measured by R-square, see Section V) for GDEM2 and AW3D30 global DEMs. Therefore, in what follows isotropic Gaussian covariation function is used for DEM error.

The DEM error properties accepted in the mvcNI+fBm can be summarized as follows: normal distribution, stationary fine-scale error with Gaussian isotropic spatial correlation function, uncorrelated with the error-free DEM. The goal of the mvcNI+fBm estimator is to evaluate types and coefficients of regression functions $\sigma_e^2(\mathbf{p})$ and $\sigma_{Corr}^2(\mathbf{p})$ using large number of $\hat{Z}_{grid}$ patches.

The error-free DEM is modeled as a fractal surface, namely non-stationary 2D fractal Brownian motion (fBm) [49]. The fBm model is suitable for describing the Earth surface that was shown to fulfill the multiscale, fractal properties [45, 50, 51]. According to fBm model, the error-free DEM increments $\Delta Z(t,s) = Z(t,s) - Z(0,0)$ with respect to the patch central pixel $Z(0,0)$ represent a non-stationary random Gaussian process with covariation matrix

$$R_Z(t_1,s_1,t_2,s_2) = \langle \Delta Z(t_1,s_1) \cdot \Delta Z(t_2,s_2) \rangle =$$
$$= 0.5\sigma_x^2 \left( \left(t_1^2 + s_1^2\right)^{H_q} + \left(t_2^2 + s_2^2\right)^{H_q} - \left((t_1-t_2)^2 + (s_1-s_2)^2\right)^{H_q} \right).$$

where $H_q \in (0,1)$ is the Hurst exponent describing fBm-field roughness ($H_q \to 0$ relates to a rougher terrain, $H_q \to 1$ to a smoother one), $\sigma_x$ describes fBm amplitude as SD of elevation increments on unit distance. The covariation matrix of measured increments $\Delta \hat{Z}(t,s) = \hat{Z}(t,s) - \hat{Z}(0,0)$ can be shown to have the following form:

$$R_{\hat{Z}}(t_1,s_1,t_2,s_2) = \langle \Delta \hat{Z}(t_1,s_1) \cdot \Delta \hat{Z}(t_2,s_2) \rangle = R_Z(t_1,s_1,t_2,s_2) + $$
$$+ R_e(t_1-t_2, s_1-s_2) + R_e(0,0) - R_e(t_1,s_1) - R_e(t_2,s_2) \quad . \quad (3)$$

*B. Bivariate model of measurement error variance and correlation*

Among the multitude of factors influencing the DEM error, in this paper we have considered only two instrument-induced factors valid for photogrammetric DEMs: the stacking number and epipolar line error.

Stacking is aggregation of data obtained from multiple passes of the instrument. Stacking number, $N_{stk}$, is number of image stereo pairs used to estimate elevation. It is selected here as the first predictor of DEM error: $p_1 = N_{stk}$. Conceptually, measurement error variance should decrease as $N_{stk}^{-1}$ due to averaging over stack of DEM images. According to this model, $\sigma_e^2$ should converge to zero with $N_{stk}$ approaching infinity, which is not realistic due to finite correlation kernel (window) size [52]. Therefore, we consider more adequate model $\sigma_e^2 = 1 + N_{stk}^{-1}$ (in what follows, we represent regression models in simplified form by setting model coefficients to unity).

In stereovision and photogrammetry, finding elevation of a terrain object consists in 1D search of the object image on the left(right) image along the corresponding epipolar line at the right(left) image. The epipolar line on one stereo pair image is formed by projecting all 3D points having the same projection in the other stereo pair image [53]. The object displacement (w.r.t. a reference elevation, typically $Z = 0$) along the epipolar line is called disparity $D$, which is related to elevation as $Z = D(r/B)$, where $B$ denotes the base-to-height ratio, $r$ is a sensor spatial resolution [54]. In Fig. 3, elevation $Z = 0$ corresponds to point A and line A-B is the corresponding epipolar line. Object E has higher elevation (and disparity) than object C and is placed farer from point A along the epipolar line. Position of the reference point A and orientation of the epipolar line are determined from the instrument calibration data and are subject to errors. The deviation of the estimated epipolar line from the true one is called epipolar line error [54]. It has been shown that camera errors propagate to epipolar line error in a way dependent on disparity [55]. As a result, standard deviation of distance between points on the true and estimated epipolar lines is increasing with disparity: in average $\|A-A'\| < \|C-C'\| < \|D-D'\| < \|B-B'\|$. The object disparity is found by maximizing correlation between left and right stereo images within the search window moving along the epipolar line. As estimated epipolar line deviates from its true position, the value of correlation drops, correlation profile maximum smears, and disparity estimation accuracy decreases. In this manner, $\sigma_e^2$ becomes dependent on $D$, and consequently on elevation $Z$. We, therefore, select elevation $Z$ as the second predictor of DEM error: $p_2 = Z$.

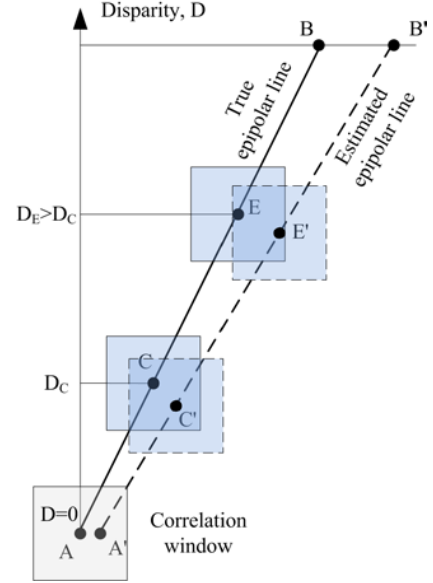

Fig. 3. Illustration of epipolar line error.

Obviously, the influence of epipolar line error depends on the accuracy of onboard calibration data. In the experimental part of the paper, we show that ASTER GDEM2 with calibration data of relatively low accuracy [54] demonstrate strong influence of elevation on DEM error. On the contrary, for AW3D and AW3D30 DEMs with calibration data of relatively high accuracy [56], this dependence is not observed.

In the absence of exact model of epipolar line error influence on $\sigma_e^2$, we propose to model it in the form $1 + Z^m$, where $m$ takes values 1 or 2 in our study (formalizing linear and quadratic dependence hypotheses of measurement error parameters on elevation).

Combining the two univariate terms $1 + N_{stk}^{-1}$ and $1 + Z^m$, we get for $\sigma_e^2$ a bivariate model in the form $(1+Z^m) + (1+Z^m)N_{stk}^{-1} = 1 + N_{stk}^{-1} + Z^m + Z^m N_{stk}^{-1}$, leading to the following expression:

$$\sigma_e^2(\mathbf{p}) = \sigma_{\text{Var}01}^2 + \sigma_{\text{Var}02}^2 N_{\text{stk}}^{-1} + c_{\text{VarZ}1}Z^m + c_{\text{VarZ}2}Z^m N_{\text{stk}}^{-1}, \quad (4)$$

where $\mathbf{p} = (N_{\text{stk}}, Z)$, $N_p = 2$ is two-dimensional predictor vector.

In (4), $\sigma_{\text{Var}01}^2 + c_{\text{VarZ}1}Z^m$ component is shared by all DEM realizations in the stack and related to the underlying surface properties. This error component can be viewed as measurement error of DEM obtained from noise-free stereo images. The term $\sigma_{\text{Var}02}^2 + c_{\text{VarZ}2}Z^m$ is the variance of component that is uncorrelated between DEM realizations in the stack (due to different sensor noise realizations in each stereo pair and different viewing geometries).

For photogrammetric DEM, measurement error spatial correlation width is related to the correlation peak sharpness and should not exceed correlation window size. The value of $\sigma_{\text{Corr}}^2$ is supposed not to change from one stereo pair to another one. The averaging over stereo pairs does not affect $\sigma_{\text{Corr}}^2$ of random components and could slightly increase it as a result of spatial smoothing due to horizontal registration error of stereo pairs. Therefore, $\sigma_{\text{Corr}}^2$ could slightly increase with $N_{\text{stk}}$. As epipolar line error smears correlation peak, $\sigma_{\text{Corr}}^2$ is expected to increase with elevation.

We propose to use the same model for $\sigma_{\text{Corr}}^2$ as for $\sigma_e^2$:

$$\sigma_{\text{Corr}}^2(\mathbf{p}) = \sigma_{\text{Corr}01}^2 + \sigma_{\text{Corr}02}^2 N_{\text{stk}}^{-1} + c_{\text{CorrZ}1}Z^m + c_{\text{CorrZ}2}Z^m N_{\text{stk}}^{-1}. \quad (5)$$

In addition to full models (4) and (5), we have also considered two simpler submodels for both $\sigma_e^2$ and $\sigma_{\text{Corr}}^2$. The first submodel uses only predictor $N_{\text{stk}}$, and the second one employs only $Z$. Taking into account two possible values for $m$ (1 or 2), we obtain a set of six alternative models to test for the DEM measurement error parameters: $1$, $1 + N_{\text{stk}}^{-1}$, $1 + Z$, $1 + Z^2$, $1 + N_{\text{stk}}^{-1} + Z + ZN_{\text{stk}}^{-1}$, $1 + N_{\text{stk}}^{-1} + Z^2 + Z^2 N_{\text{stk}}^{-1}$. In what follows, we use the mvcNI+fBm both to select the most relevant model for each considered DEM and estimate the corresponding model coefficients.

## IV. NO-REFERENCE ESTIMATION OF MEASUREMENT ERROR PARAMETERS: THE MVCNI+FBM ESTIMATOR

### A. Generalized description of the adapted mvcNI+fBm estimator of DEM error parameters

The main processing stages of the mvcNI+fBm method include (1) initialization, (2) patches homogeneity estimation, (3) grouping of image patches, (4) processing of the formed groups, and (5) multivariate estimation of context-dependent error parameters (see Fig. 4) [26]. In the mvcNI+fBm method, homogeneity estimation stage depends on DEM error parameters; therefore, iterative repeating of stages 2-5 is required. Originally the mvcNI+fBm was proposed for vector remote sensing images. In this paper, it is adapted to the case of gridded DEM images that represents single-channel data. Let us consider each stage of the mvcNI+fBm in more detail.

To estimate each error parameter, one specific call of sequence of stages (3)-(5) is performed. During each call, one parameter is fixed while the other ones are estimated, and vice-versa. For the purpose of clarity, we describe below the proposed estimation procedure for an arbitrary error parameter $s$, meaning either noise SD, $s = \sigma_e$, or spatial correlation width, $s = \sigma_{\text{Corr}}$.

At the initialization stage, the image to process is split into non-overlapping patches. Each patch represents a group of adjacent pixels of size $N$ by $N$ pixels. The patches with known unreliable data are removed from further consideration. This stage is application-specific and is described in the next experimental section. Overall number of retained patches is referred to as $N_{\text{pt}}$.

In the mvcNI+fBm, homogeneity index is measured as a ratio between lower bound on parameter $s$ estimation SD (Cramér–Rao lower bound, CRLB), $\sigma_s$, and parameter value: $r_{\text{HA}} = \sigma_s / s$. For homogeneous patches, it takes low values. For example, for standard deviation parameter, very homogeneous patches represents pure DEM error, $\sigma_{\sigma_e^2}$ approaches to SD of sample variance: $\sigma_{\sigma_e^2} \approx \sigma_e \sqrt{2/(N^2-1)}$, and $r_{\text{HA}} \approx \sqrt{2/(N^2-1)}$. For $N = 11$ used in the experimental part, $\sqrt{2/(N^2-1)} = 0.129$. For heterogeneous patches, actual DEM and DEM error cannot be separated, $\sigma_s$ increases as well as $r_{HA}$ ratio. In extreme cases, for very rough patches $r_{HA} \to \infty$. The value of $r_{HA}$ indicates how much information about error parameter $s$ can be extracted from a given patch with patch homogeneity increasing as $r_{HA}$ decreasing. Correspondingly, in the mvcNI+fBm, homogeneous areas with low $r_{HA}$ index were called Noise Informative (NI). Homogeneity estimation stage involves estimation of error-free DEM parameters, value of $\sigma_s$, and, finally, value of $r_{HA}$ for each patch (see the next subsection for details). Notice that a specific value of $r_{HA}$ is defined for each considered error parameter. Therefore, homogeneity estimation stage (and subsequent stages of mvcNI+fBm) is repeated for each error model parameter.

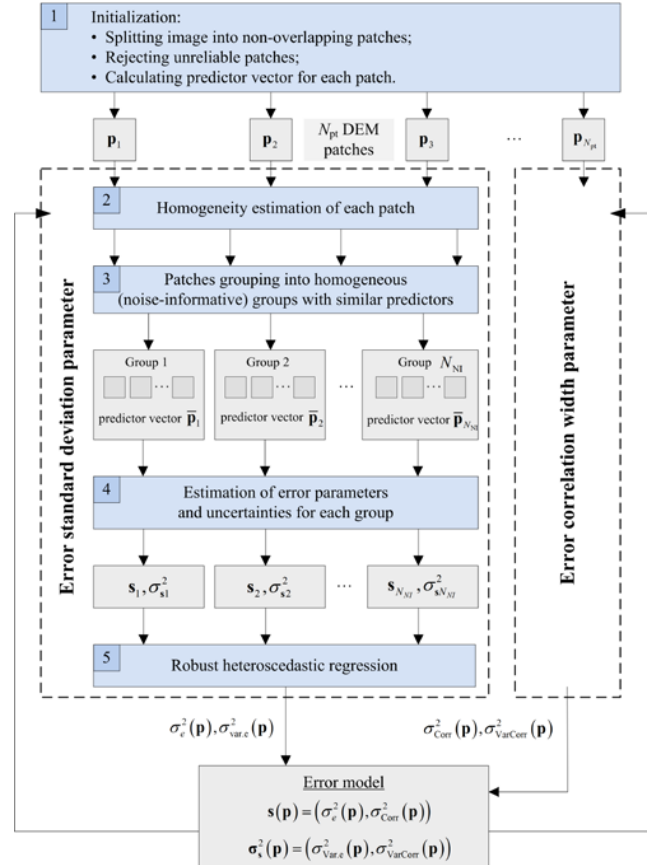

Fig. 4. Processing pipeline of the mvcNI+fBm method

Let us consider two patches with the same error parameter $s$ and CRLBs $\sigma_{s1}$ and $\sigma_{s2}$. The two patches considered together provide more information on $s$ than each of them considered individually. By joint processing of two patches CRLB value could be potentially reduced to $\sigma_s = \dfrac{\sigma_{s2}\sigma_{s1}}{\sigma_{s2}+\sigma_{s1}} < \min(\sigma_{s2},\sigma_{s1})$. Therefore, for two patches $r_{HA} < \min(r_{HA1}, r_{HA2})$. From this point of view, several patches considered together are "more homogeneous" and more "noise informative" than each of them. For a group of $N_{group}$ patches with similar DEM properties, homogeneity index decreases approximately as $1/\sqrt{N_{group}}$. The mvcNI+fBm utilizes this property by grouping moderately heterogeneous and homogeneous patches with similar predictor values into even more homogeneous groups. Each group is further processed as a whole. In this manner, even moderately heterogeneous patches could form a homogeneous group, thus relaxing requirement to actual DEM variability and increasing number of individual error parameter estimates.

The grouping stage splits all available patches into cells with similar predictor values (implying similar error properties). Similarity of predictors is measured as $d(\mathbf{p}_1,\mathbf{p}_2) = \sqrt{\sum_{j=1}^{N_p} w_j^2 \cdot (\mathbf{p}_1(j) - \mathbf{p}_2(j))^2}$ where $w_j$ is the associated weight (to guarantee the desired resolution with respect to predictors). Weights $w_j$, $j=1...N_p$, serve two interdependent goals: the first one is to limit the error parameter variation within the group and the second one is to assure the desired resolution with respect to predictors. To provide resolution $\Delta p_j$ w.r.t j-th predictor, the weight should be set as $w_j = \Delta p_j^{-1}$ (the used weight values for the DEM error parameter estimation are specified in the experimental part of the paper). In each cell, patches are sorted in ascending order of $r_{HA}$ value (from the most homogeneous to the most heterogeneous ones). Patches are sequentially assigned to groups. The groups are considered homogeneous when $r_{HA}$ drops below threshold value of 0.15 and $N_{group} < 15$ (a larger size of an homogeneous group means that the set of patches in the group is more heterogeneous and consequently extracting error parameters from such a group may become less reliable; in the tests carried out later in the experimental part of the paper, we found 15 to be a reasonable group size). We define by $N_{NI}$ the number of the found NI groups. For i-th group, mean value of predictor vector $\bar{\mathbf{p}}_i$ is calculated.

At the next stage (4) of processing the formed groups, for an i-th group a single error parameter $s_i$ ($\sigma_{e,i}^2$ or $\sigma_{Corr,i}^2$) is estimated using all patches in this group. Thus, each group provides one observation $\hat{s}_i$ of the error parameter $s_i$ for a given value of predictor vector $\bar{\mathbf{p}}_i$. Note that the lower bound on $\hat{s}_i$ estimation error, denoted as $\sigma_{s_i}^2$, is derived as an additional output of the estimation procedure (see the next subsection).

The final stage (5) of estimation of multivariate context-dependent error parameter $s$ mathematically corresponds to the associated robust, multivariate, heteroscedastic regression problem so as to estimate $s$ as a function of its predictors [26]. Multivariate linear regression (nonlinear regression can be applied in a similar way) is thus used in the mvcNI+fBm rather than simple univariate linear regression typically considered in the past for signal-dependent noise parameter estimation. Robustness is needed to cope with gross DEM errors (outliers, blunders) [5].

B. *Mathematical details of the mvcNI+fBm estimator*

Let us formally introduce the mvcNI+fBm estimator. Using the definitions of Subsection 3.1, the log-likelihood function of image intensities within a single patch is given by

$$\ln L(\Delta \hat{\mathbf{Z}}; \boldsymbol{\theta}) = -\frac{1}{2}\left[ \Delta \hat{\mathbf{Z}}^T \mathbf{R}_{\Delta \hat{\mathbf{Z}}}^{-1} \Delta \hat{\mathbf{Z}} + \ln|\mathbf{R}_{\Delta \hat{\mathbf{Z}}}| \right], \quad (6)$$

where $\Delta \hat{\mathbf{Z}}$ is a $N^2 \times 1$ sample composed of all $\Delta \hat{Z}(t,s)$ within the patch, $k$-th $\Delta \hat{\mathbf{Z}}$ element and patch pixel with coordinates $(t_k, s_k)$ are related to each other by $k = t_k + N s_k + N_h(N+1)+1$, $\boldsymbol{\theta}=(\sigma_x^2, H_q, \sigma_e^2, \sigma_{Corr}^2)$ denotes the full parameter vector characterizing both error-free DEM and DEM error.

The information contained in the sample $\Delta \hat{\mathbf{Z}}$ is characterized by the Fisher Information Matrix (FIM) $\mathbf{I}_\theta$ with elements $I_{\theta_i \theta_j} = \dfrac{1}{2} tr\left( \mathbf{R}_{\Delta \hat{\mathbf{Z}}}^{-1} \dfrac{\partial \mathbf{R}_{\Delta \hat{\mathbf{Z}}}}{\partial \theta_i} \mathbf{R}_{\Delta \hat{\mathbf{Z}}}^{-1} \dfrac{\partial \mathbf{R}_{\Delta \hat{\mathbf{Z}}}}{\partial \theta_j} \right)$, $i,j=1...4$ [57].

Before estimating error parameters, we first need to determine error-free DEM parameters, $\sigma_x^2$ and $H_q$, using the following estimator:

$$\left[\hat{\sigma}_x^2, \hat{H}_q\right] = \arg\min_{\sigma_x^2 \geq 0,\ 0 < H_q < 1, \sigma_e^2 = \hat{\sigma}_e^2, \sigma_{Corr}^2 = \hat{\sigma}_{Corr}^2} \left[\ln L(\Delta \hat{\mathbf{Z}}; \boldsymbol{\theta})\right]. \quad (7)$$

In (7), we specify that error parameters are fixed at this stage at their current estimated values $\hat{\sigma}_e^2$ and $\hat{\sigma}_{Corr}^2$. Error-free DEM parameter estimates are only reliable for patches with ratio $\sigma_x^2/\sigma_e^2$ that can be understood as local signal-to-noise ratio (SNR) within the given DEM patch. High SNR patches with $\sigma_x^2 / \sigma_e^2 > 2$ are called Texture Informative (TI). For other patches, Hurst exponent estimation is not reliable and is evaluated using the inverse distance interpolation method [58] based on reliable estimates $\hat{H}_q$ from TI patches. In this manner, the interpolated Hurst exponent values $\hat{H}_{q.\text{interp}}$ are obtained for each patch. To characterize interpolation error, we sequentially exclude one TI patch and interpolate Hurst exponent value for this patch using remaining TI patches. The interpolation error SD $\bar{\sigma}_{H_{q.\text{interp}}}$ is calculated as SD of difference $\hat{H}_{q.\text{interp}} - \hat{H}_q$ for all TI patches.

Two estimators for the error variance $\sigma_e^2$ and spatial correlation width $\sigma_{\text{Corr}}^2$ can now be defined in the following form:

$$\left[\hat{\sigma}_x^2, \hat{H}_q, \hat{\sigma}_e^2\right] = \underset{\sigma_x^2 \geq 0,\ 0 < H_q < 1,\ \sigma_e^2 \geq 0,\ \sigma_{\text{Corr}}^2 = \hat{\sigma}_{\text{Corr}}^2}{\arg\min} \left[\ln L(\Delta\hat{\mathbf{Z}};\boldsymbol{\theta}) - \frac{\left(H_q - \hat{H}_{q.\text{interp}}\right)^2}{2\bar{\sigma}_{H_{q.\text{interp}}}^2}\right], \quad (8)$$

$$\left[\hat{\sigma}_x^2, \hat{H}_q, \hat{\sigma}_{\text{Corr}}^2\right] = \underset{\sigma_x^2 \geq 0,\ 0 < H_q < 1,\ \sigma_e^2 = \hat{\sigma}_e^2,\ \sigma_{\text{Corr}}^2 \geq 0}{\arg\min} \left[\ln L(\Delta\hat{\mathbf{Z}};\boldsymbol{\theta}) - \frac{\left(H_q - \hat{H}_{q.\text{interp}}\right)^2}{2\bar{\sigma}_{H_{q.\text{interp}}}^2}\right]. \quad (9)$$

In (8) and (9), we have taken into account the prior on Hurst exponent as the additional term in the optimized log-likelihood function. This prior has maximal effect for the very homogeneous patches with low SNR, for which reliable Hurst exponent estimation is impossible. The effect becomes less pronounced with SNR increase.

FIMs $\mathbf{I}_{(\sigma_x^2, H_q, \sigma_e^2)}$ and $\mathbf{I}_{(\sigma_x^2, H_q, \sigma_{\text{Corr}}^2)}$ for the estimators (8) and (9) respectively are obtained from general matrix $\mathbf{I}_\theta$ by eliminating row and column corresponding to the missing parameter and adding prior information on $H_q$ by substituting $I_{H_q H_q}$ component with $I_{H_q H_q} + \bar{\sigma}_{H_{q.\text{interp}}}^{-2}$.

Cramer-Rao lower bounds on $\sigma_e^2$ and $\sigma_{\text{Corr}}^2$ estimate error are obtained by inverting $\mathbf{I}_{(\sigma_x^2, H_q, \sigma_e^2)}$ and $\mathbf{I}_{(\sigma_x^2, H_q, \sigma_{\text{Corr}}^2)}$. The last elements of these matrices, $\sigma_{\sigma_e^2}^2$ and $\sigma_{\sigma_{\text{Corr}}^2}^2$, are CRLBs on $\sigma_e^2$ and $\sigma_{\text{Corr}}^2$ estimate errors, respectively.

Taking into account error-free DEM and error independence between patches, estimators for the group of $N_{\text{group}}$ patches can be straightforwardly formulated using (8) and (9):

$$\left[\hat{\mathbf{D}}_x, \hat{\mathbf{H}}_q, \hat{\sigma}_e^2\right] = \underset{\mathbf{D}_x \geq 0,\ 0 < \mathbf{H}_q < 1,\ \sigma_e^2 \geq 0,\ \sigma_{\text{Corr}}^2 = \hat{\sigma}_{\text{Corr}}^2}{\arg\min} \left[\sum_{v=1}^{N_{\text{group}}} \left(\ln L(\Delta\hat{\mathbf{Z}}_v;\boldsymbol{\theta}_v) - \frac{\left(H_{q.v} - \hat{H}_{q.\text{interp}.v}\right)^2}{2\bar{\sigma}_{H_{q.\text{interp}}}^2}\right)\right], (10)$$

$$\left[\hat{\mathbf{D}}_x, \hat{\mathbf{H}}_q, \hat{\sigma}_{\text{Corr}}^2\right] = \underset{\mathbf{D}_x \geq 0,\ 0 < \mathbf{H}_q < 1,\ \sigma_e^2 = \hat{\sigma}_e^2,\ \sigma_{\text{Corr}}^2 \geq 0}{\arg\min} \left[\sum_{v=1}^{N_{\text{group}}} \left(\ln L(\Delta\hat{\mathbf{Z}}_v;\boldsymbol{\theta}_v) - \frac{\left(H_{q.v} - \hat{H}_{q.\text{interp}.v}\right)^2}{2\bar{\sigma}_{H_{q.\text{interp}}}^2}\right)\right], (11)$$

where $\Delta\hat{\mathbf{Z}}_v$ is the sample for $v$-th patch, $\boldsymbol{\theta}_v = (\sigma_{x.v}^2, H_{q.v}, \sigma_e^2, \sigma_{\text{Corr}}^2)$, $\sigma_{x.v}^2$, $H_{q.v}$, and $\hat{H}_{q.\text{interp}.v}$ are $\sigma_x^2$, $H_q$, and $\hat{H}_{q.\text{interp}}$ for $v$-th patch, $\mathbf{D}_x = \left(\sigma_{x.1}^2, \sigma_{x.v}^2, ..., \sigma_{x.N_{gr}}^2\right)$, $\mathbf{H}_q = \left(H_{q.1}, H_{q.2}, ..., H_{q.N_{gr}}\right)$. Note that the same error parameters are assumed for all patches in the group.

For non-overlapping patches, error parameter estimates are independent. Therefore, CRLBs on $\sigma_e^2$ and $\sigma_{\text{Corr}}^2$ using a group of patches are given by $\sigma_{\sigma_e^2}^2 = 1/\sum_{v=1}^{N_{\text{group}}} \left(\sigma_{\sigma_{e.v}^2}^{-2}\right)$ and $\sigma_{\sigma_{\text{Corr}}^2}^2 = 1/\sum_{v=1}^{N_{\text{group}}} \left(\sigma_{\sigma_{\text{Corr}.v}^2}^{-2}\right)$, respectively.

Finally, let us provide $\mathbf{R}_{\Delta\hat{\mathbf{Z}}}$ derivatives with respect to $\theta_i$, $i = 1...4$:

$$\frac{\partial \mathbf{R}_{\Delta\hat{\mathbf{Z}}}(k_1, k_2)}{\partial \sigma_x^2} = \sigma_x^{-2} R_{\Delta\mathbf{Z}}(t_{k_1}, s_{k_1}, t_{k_2}, s_{k_2}), \quad (12)$$

$$\frac{\partial \mathbf{R}_{\Delta\hat{\mathbf{Z}}}(k_1, k_2)}{\partial H_q} = 0.5\sigma_x^2 \left(\log(t_{k_1}^2 + s_{k_1}^2)(t_{k_1}^2 + s_{k_1}^2)^{H_q} + \\ + \log(t_{k_2}^2 + s_{k_2}^2)(t_{k_2}^2 + s_{k_2}^2)^{H_q} - \\ - \log((t_{k_1} - t_{k_2})^2 + (s_{k_1} - s_{k_2})^2)((t_{k_1} - t_{k_2})^2 + (s_{k_1} - s_{k_2})^2)^{H_q}\right), \quad (13)$$

$$\frac{\partial \mathbf{R}_{\Delta\hat{\mathbf{Z}}}(k_1, k_2)}{\partial \sigma_e^2} = \begin{pmatrix} 1 + \exp\left(-\frac{d_0^2}{2\sigma_{\text{Corr}}^2}\right) - \\ -\exp\left(-\frac{d_1^2}{2\sigma_{\text{Corr}}^2}\right) - \exp\left(-\frac{d_2^2}{2\sigma_{\text{Corr}}^2}\right) \end{pmatrix}, \quad (14)$$

$$\frac{\partial \mathbf{R}_{\Delta\hat{\mathbf{Z}}}(k_1, k_2)}{\partial \sigma_{\text{Corr}}^2} = \frac{\sigma_e^2}{2\sigma_{\text{Corr}}^4} \begin{pmatrix} d_0^2 \exp\left(-\frac{d_0^2}{2\sigma_{\text{Corr}}^2}\right) - \\ -d_1^2 \exp\left(-\frac{d_1^2}{2\sigma_{\text{Corr}}^2}\right) - d_2^2 \exp\left(-\frac{d_2^2}{2\sigma_{\text{Corr}}^2}\right) \end{pmatrix}, (15)$$

where $d_0 = \sqrt{(t_{k_1} - t_{k_2})^2 + (s_{k_1} - s_{k_2})^2}$, $d_1 = \sqrt{t_{k_1}^2 + s_{k_1}^2}$, $d_2 = \sqrt{t_{k_2}^2 + s_{k_2}^2}$.

## V. APPLICATION OF ADAPTED MVCNI+FBM ESTIMATOR TO REAL DEMS ERROR ANALYSIS

In this section, we apply the proposed mvcNI+fBm method to two global DEMs with 1 arcsec (30 m at equator) spatial resolution: GDEM2 and AW3D30. More attention is paid to GDEM2 as it reveals the most complex measurement error behavior. To prove the mvcNI+fBm scalability, we test it on a sample granule of AW3D DEM with 5 m spatial resolution. The obtained results for each DEM are compared with respective DEM accuracy analysis available in the literature.

## A. Experiment settings

For GDEM2 and AW3D30 DEMs, 59 tiles were used to estimate $\sigma_e^2$ and $\sigma_{Corr}^2$ model coefficients with the following lower-left (southwest) corner pixels: N27-28E086-087, N30-32E035, N33-35E076, N47-48W001-002, N48-49E001-003, N48-49E031-032, N49-50E036-037, N50E099, N50E103, N51-52E099, N52E101-102, N53E103, S23-28W067-070. Here, hyphen defines range of tiles with respect to latitude, longitude or both; N and S denote north and south latitude, respectively; W and E denote west and east longitude, respectively.

Each tile has fixed latitudinal and longitudinal spatial resolution of 1 arc-second (approximately 30 m at the equator) but expressed in meters, resolution with respect to image rows and columns differs. To compensate for this effect, tiles were resampled to coarser 90 m resolution cell with respect to both spatial coordinates. For each tile, 10000 non-overlapping patches of 11 by 11 pixels were selected on a regular grid. The selected size of the patch is an experimentally found compromise between accuracy of error parameters estimates (that increase with patch size) and DEM patch homogeneity (that decreases with patch size). The stacking number for each pixel of these patches was obtained using quality assessment (QA) files supplied for each tile (the QA file has the same number of rows and columns as DEM tile with each pixel representing the corresponding $N_{stk}$ value. QA file has extension .num for GDEM2 and suffix "STK" for ALOS World 3D). QA files were also interpolated to 90m grid to match DEM tiles.

We set two criteria to decide whether a patch is reliable for measurement error parameter estimation: (1) it contains data provided by the "normal" workflow for the particular DEM; (2) patch is not significantly affected by GDEM2 stacking procedure artifacts (see the next paragraph). The first criterion for photogrammetric DEM excludes pixels where disparity measurement procedure fails, primary from lack of data (e.g. caused by clouds), water body (that are typically masked out) and low correlation areas (e.g. deserts). Those pixel are substituted from different sources (e.g. other DEMs) or remain as voids and do not characterize measurement error. Such pixels can be identified by a negative stacking number for GDEM2 and from mask information file (marked by "MSK" suffix) for ALOS World 3D.

The second criterion is for handling another problem causing gross elevation error is related to GDEM2 stacking procedure: it produces false elevation discontinuities if stacking number changes severely. This problem is illustrated in Fig. 5, where a GDEM2 DEM patch, the corresponding stacking number map, and the AW3D30 patch from the same location are shown. The false elevation discontinuity correlated with stacking number discontinuity is clearly visible, while at AW3D30 patch this feature is missing. To avoid this source of gross errors, the minimum $N_{stk.min}$ and maximum $N_{stk.max}$ numbers of stereo pairs were calculated for each patch. The patch was rejected if $N_{stk.max} / N_{stk.min} > 2$ thus limiting stacking number variability. All patches that were found reliable are further processed by the

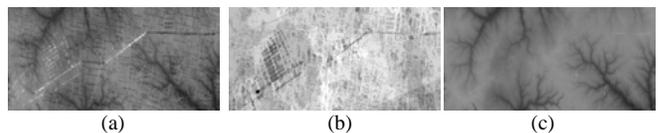

| (a) | (b) | (c) |

Fig. 5. Illustration of GDEM2 gross errors related to stacking procedure. GDEM2 patch (a), GDEM2 stacking number (b), AW3D30 patch (c). Black color corresponds to elevation of 65m and $N_{stk} = 0$, white color to 290m and $N_{stk} = 11$.

mvcNI+fBm estimator.

Predictor weights were set as $w_1 = 1$ and $w_2 = 0.01 m^{-1}$ for $N_{stk}$ and $Z$, respectively. Required homogeneity index for NI groups was experimentally set to 0.125. We found that in all cases mvcNI+fBm converged in less than 15 iterations.

We use classical tools of linear regression to characterize regression models: generalized determination coefficient, partial residual plot, significance test for predictor values using t-statistic [59, 60]. The generalized determination coefficient [61]: $R^2 = 1 - \exp\left(-\frac{2}{N_{NI}}(l_R - l_U)\right)$, where $l_R$ is the log-likelihood for regression model restricted only with intercept, $l_u$ is the log-likelihood of the unrestricted regression model, $N_{NI}$ is the number of estimates (number of NI groups). Assuming normal distribution of error parameter estimate errors provided by the mvcNI+fBm and omitting constants shared by both $l_R$ and $l_u$, the log-likelihoods for restricted and unrestricted models take the following quite simple form:

$$l_R = -\frac{1}{2}\sum_{i=1}^{N_{NI}} \frac{(\hat{s}_i - s_{const})^2}{\sigma_{s_i}^2} \text{ and } l_u = -\frac{1}{2}\sum_{i=1}^{N_{NI}} \frac{(\hat{s}_i - s_{pr.i})^2}{\sigma_{s_i}^2}, \text{ where } s_{pr.i}$$

is parameter prediction for ith measurement.

## B. ASTER GDEM2 experiment

The Advanced Spaceborne Thermal Emission and Reflection Radiometer (ASTER) Global Digital Elevation Model version 2 was released by the National Aeronautic and Space Administration (NASA), USA, and the Ministry of Economy, Trade, and Industry (METI) of Japan in 2011 [21, 22]. Experimental assessment of the GDEM2 has been published in numerous papers including [20, 21, 62] among others.

The ASTER GDEM2 was produced from data collected by the ASTER sensor on board of the NASA Terra spacecraft, which is capable of collecting in-track stereo pairs using nadir- and aft- looking near infrared cameras [63, 64]. The procedure for the GDEM2 generation is described in [54]. According to this procedure, elevation estimates are obtained via correlation-based registration with 5 by 5 pixel window. Curve fitting approach [65] is used to reach subpixel registration accuracy. The search is performed in the along-track direction neglecting a small cross-track shift component appearing due to contribution of Earth rotation effects during the stereoscopic observation period. Multiple stereo pairs (stacking number up to 50) are used to obtain elevation estimates for a given area.

**Measurement error variance analysis.** The results obtained for error parameter $\sigma_e^2$ are presented in Table 1. The first observation is that models with quadratic dependence on elevation are more significant according to $R^2$ coefficient ($R^2$ is 0.4335 and 0.6298 for models $1+Z$ and $1+Z^2$; 0.7717 and 0.8954 for models $1+N_{stk}^{-1}+Z+ZN_{stk}^{-1}$ and $1+N_{stk}^{-1}+Z^2+Z^2N_{stk}^{-1}$, respectively). Using only $N_{stk}$ predictor, we get a low determination coefficient $R^2=0.1221$ (model $1+N_{stk}^{-1}$). The second $Z$ predictor notably increases $R^2$ to 0.6298 (model $1+Z^2$). Using all terms in the model (4) increases $R^2$ even further to the value of 0.8954. Therefore, the influence of both predictors on $\sigma_e^2$ is significant. The estimated model for the GDEM2 elevation measurement error variance is given as:

$$\sigma_e^2 = 1.0293\text{m}^2 + 25.6667\text{m}^2 N_{stk}^{-1} + \\ + 4.8991\cdot 10^{-7} Z^2 + 6.1069\cdot 10^{-6} Z^2 N_{stk}^{-1} \quad (16)$$

The partial residual plots for components of the model (16) with the highest $R^2$ (Fig. 6) reveal significance of each component. Note the practical absence of outliers among mvcNI+fBm estimates (outliers percentage is about 0.1% for $\sigma_e^2$).

Let us validate further the selected model (16) both qualitatively and quantitatively in comparison with GDEM2 analysis results provided in the available literature. For the ASTER instrument, $B=0.6$ and $r=15$ m. The ratio $(r/B)$ = 15 m/pixel is conversion ratio between pixel and meter units. It is systematically used in the analysis below where conversion between measurement units is involved.

The value of $\sigma_e$ at the sea level ($H=0$) and for one stereopair ($N_{stk}=1$) is given as $\sqrt{\sigma_{\text{Var}01}^2+\sigma_{\text{Var}02}^2}=5.1668$ m. This value was estimated by Fujisada et al. in [54] (see Fig. 10, kernel size 5 by 5 pixels) to vary from 0.2 to 0.3 pixels or, equivalently, from 5 to 7.5 m, depending on the terrain. One can conclude and underline good agreement between these two estimates.

The value of $\sigma_e$ introduced in the present paper corresponds to the standard deviation of elevation bias in [20]. Therefore, we can directly compare model (16) with the results by K. Becek. The reported results of the dependence of standard deviation of bias on $N_{stk}$ make this comparison even more informative. The majority of 96 runways considered in [20] are situated at low altitudes near the sea level with the mean elevation of 234.7 m. We simplify model (16) by substituting $Z=234.7$ m getting the reduced model $\sigma_e^2=1.0563+26.0031 N_{stk}^{-1}$. The results obtained with the reduced model were compared to those shown in Fig. 10 from [21]. Both estimates are shown in Fig. 7 for $N_{stk}$ varying from 0 to 50.

The results of the mvcNI+fBm and runway methods are highly consistent for $N_{stk}>20$. For $N_{stk}<20$, $\sigma_e^2$ estimates obtained by the mvcNI+fBm are up to two times smaller than those provided by the runway method. The possible reason of this difference is the better robustness of mvcNI+fBm to gross GDEM2 errors and outliers that affect the GDEM2 for low

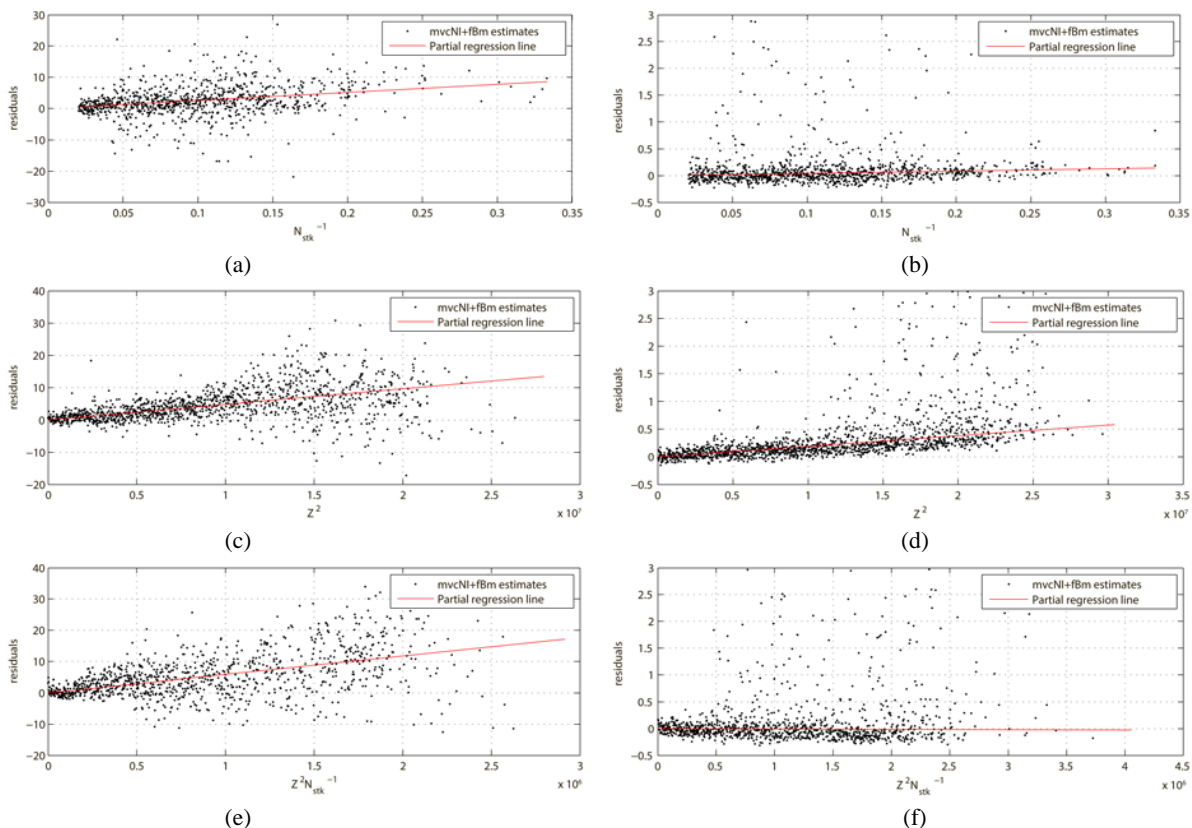

Fig. 6. Partial regression plots for $\sigma_e^2$ (a, c, e) and $\sigma_{\text{Corr}}^2$ (b, d, f) model coefficients.

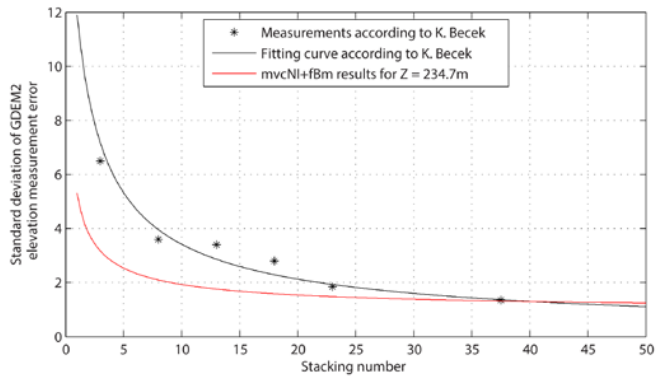

Fig. 7. Comparison of the mvNI+fBm results for the GDEM2 with the results obtained by K. Becek [22]

stacking numbers as indicated by K. Becek.

The estimates of $c_{VarZ1}$ and $c_{VarZ2}$ related to epipolar line error show very significant influence of elevation on $\sigma_e^2$: $\sigma_e$ at Z = 6000 m increases by 1.9…3.4 times as compared to this value at Z=0 m making elevation influence the predominant error source at high elevations. The epipolar line error SD is measured in [54] to be about 0.15 pixels or 3.75 m for elevation about 4000 m (Mt. Elbert). According to model (16), $\sigma_e^2$ at elevation Z=4000 m and typical number of stereo pairs $N_{stk} = 10$ increases by $4.899 \cdot 10^{-7}(4000m)^2 + 6.107 \cdot 10^{-6}(4000m)^2/10 = 17.61 m^2$. This additional SD of $\sqrt{17.61 \text{ m}^2} = 4.2\text{m}$ is very close to the value reported in [54].

**Measurement error correlation width analysis.** For the correlation width parameter (Table 1) significant dependence on elevation can be observed (determination coefficient is about 0.85-0.9 for the models $1+Z$ and $1+Z^2$) revealing epipolar line error influence on measurement error correlation function. Similarly to what occurs for modeling $\sigma_e^2$, models depending on $Z$ are less relevant than models depending on $Z^2$ for modeling $\sigma_{Corr}^2$. Joint usage of two predictors, $N_{stk}$ and $Z$, cannot definitely confirm or discard hypothesis of $N_{stk}$ influence on correlation width: t-stats for terms dependent on $N_{stk}$ have rather small values (less than 10; see Partial regression plots in Fig. 6). Therefore, we consider model $1+Z^2$ as the most relevant one for the GDEM2 measurement error spatial correlation width (numeric values are in 90 m pixels):

$$\sigma_{Corr}^2 = 0.1937 + 1.7786 \cdot 10^{-8} Z^2 \quad (17)$$

The constant term of correlation width estimate obtained by the mvcNI+fBm method for $Z = 0$ is about 0.44 pixels at the resolution of 90 m that corresponds to 2.64 pixels at the ASTER resolution of 15 m. This value is about the half-size of 5 by 5 pixels correlation window, what is reasonable. For $Z = 5000$ m, correlation width increases almost twice to $\sigma_{Corr} = 0.8$ or 4.8 ASTER pixels. The value of correlation width estimate close to the correlation window size indicates that the measurement error correlation function shape could deviate from Gaussian one at least for high elevation values. Additional research is needed to clarify this hypothesis.

*C. ALOS experiment (5 m resolution)*

ALOS World 3D is a recent global DEM that is derived from the data generated by Panchromatic Remote-sensing Instrument for Stereo Mapping (PRISM), one of onboard sensors carried in the Advanced Land Observing Satellite (ALOS). It has high spatial resolution of 0.15 arcsec (approx. 5 m at equator) and target vertical accuracy of 5 m (RMSE). The low resolution version (30 m spacing) is generated by averaging the original one [66] and is provided free of charge. Elevation is estimated by correlation coefficient maximization according to the triplet image matching algorithm with correlation window size optimization [67]. For the PRISM instrument, $B = 0.5$ (between nadir and forward/aft-looking sensors) and $r = 2.5$ m. The ratio $(r/B) = 5$ m/pixel is conversion ratio between pixel and meter units.

Sample AW3D tile for the area of Yushan mountain, Taiwan, (central coordinates are 23°28'08.4"N 120°57'36.0"E, 4336 by 4702 pixels) were kindly provided by NTT DATA. As for the GDEM2, the AW3D tile was downscaled three times to 15 m spatial resolution. For the AW3D, experiments do not reveal significance of elevation predictor on measurement error parameters (both $\sigma_e^2$ and $\sigma_{Corr}^2$, see Table

TABLE I
THE RESULTS OF ELEVATION MEASUREMENT ERROR PARAMETER ESTIMATION OBTAINED WITH THE MVCNI+FBM FOR THE ASTER GDEM2

| Model type | $R^2$ | Parameters | Parameter SDs | t-stats |
|---|---|---|---|---|
| **Measurement noise variance** | | | | |
| $1+N_{stk}^{-1}+Z^2+Z^2 N_{stk}^{-1}$ | 0.8954 | $\sigma_{Var01}^2 = 1.0293 \text{ m}^2$ | 0.0648 m$^2$ | 15.8765 |
| | | $\sigma_{Var02}^2 = 25.6637 \text{ m}^2$ | 0.9898 m$^2$ | 25.9296 |
| | | $c_{VarZ1} = 4.8991 \cdot 10^{-7}$ | $1.6400 \cdot 10^{-8}$ | 29.8722 |
| | | $c_{VarZ2} = 6.1069 \cdot 10^{-6}$ | $2.6929 \cdot 10^{-7}$ | 22.6777 |
| **Measurement error correlation width (in 90 m pixels)** | | | | |
| $1+Z^2$ | 0.8977 | $\sigma_{Corr01}^2 = 0.1937$ | 0.0023 | 83.0768 |
| | | $c_{CorrZ1} = 1.7786 \cdot 10^{-8}$ | $2.3397 \cdot 10^{-10}$ | 76.0193 |

In total, 5331 $\hat{\sigma}_e^2$ estimates and 8044 $\hat{\sigma}_{Corr}^2$ estimates were obtained by the mvcNI+fBm. They cover predictor variation intervals from 3 to 49 for $N_{stk}$ and from -400 to about 5500 m for $Z$. The percentage of detected outliers is about 0.1% for the error variance and about 7.5% for error correlation width.

2). This result indicates negligible epipolar line error for AW3D that can be related to higher accuracy of ALOS satellite ephemeris data [56] as compared to Terra satellite [54]. Similarly to GDEM2, measurement error variance is strongly dependent on stacking number. Restricted amount of data available covers only the limited range of $N_{stk}$ change from about 4 to 7. Therefore, $\sigma_e^2$ for infinite number of $N_{stk}$ cannot be estimated with high precision. Surprisingly, $\sigma_e^2$ for AW3D and GDEM2 for sea level elevation is of the same order: $\sigma_e$ is about 5 m for $N_{stk}=1$. Converted to disparity error in sensor pixels, 5 m SD corresponds to about 0.2 pixels for GDEM2 (5 m/25(m/pixel)) and 1 pixel for AW3D (5 m/5(m/pixel)). Such an elevated error SD may be related to peculiarities of stereo-matching algorithm implementation for AW3D. Measurement error correlation width for the AW3D is about 0.39 pixels at the resolution of 15 m.

### D. ALOS experiment (30 m resolution)

For the AW3D30 as for AW3D, the experiments reveal significance of stacking number predictors for $\sigma_e^2$. For $\sigma_{Corr}^2$, both predictors are not significant. Measurement error variance for the AW3D30 is lower than the one for the GDEM2 even in the best settings (large stacking number, sea level elevation): $\sigma_{Var01} = 0.6140$ m for the AW3D30 as compared to $\sigma_{Var01} = 1.0145$ m for GDEM2. Taking into account that the AW3D30 is obtained by averaging AW3D pixels by 6 times w.r.t. both coordinates, $\sigma_e$ for the AW3D30 should be 6 times lower than for the AW3D. This is actually observed for $N_{stk}=1$: $\sigma_{e,AW3D}(1)/6 \approx 0.83$ m as compared to $\sigma_{e,AW3D30}(1) \approx 0.78$ m. For large values of $N_{stk}$, measurement error SD for the AW3D30 is comparable to the quantization error. In this extreme case, the mvcNI+fBm might not reveal decay of $\sigma_e^2$ with stacking number. Measurement error correlation width for the AW3D30 is about 0.2556 pixels at the resolution of 90 m.

### VI. CONCLUSION

In this paper, we have proposed and investigated application of the blind noise parameter estimator to characterize gridded DEM vertical error, specifically fine-scale elevation measurement error.

While BNPE is a well developed area, the new application scenario has not been covered by existing methods: the elevation measurement error SD is dependent on several predictors, while existing methods deal with signal-dependent noise model with only one predictor, namely image intensity. Therefore, the recently proposed mvcNI+fBm estimator that is able to deal with multivariate noise signal-dependency has been modified and used estimate both SD and spatial correlation width of DEM measurement error.

In the experimental part of the paper, the mvcNI+fBm has

TABLE II
THE RESULTS OF ELEVATION MEASUREMENT ERROR PARAMETER ESTIMATION OBTAINED WITH THE MVCNI+FBM FOR THE ALOS 3D WORLD, 5 M

| Model type | $R^2$ | Parameters | Parameter SDs | t-stats |
|---|---|---|---|---|
| **Measurement error variance** | | | | |
| $1+N_{stk}^{-1}$ | 0.27306 | $\sigma_{Var01}^2 = 0.3558$ m$^2$ | 0.1441 m$^2$ | 2.4680 |
| | | $\sigma_{Var02}^2 = 24.2965$ m$^2$ | 0.7402 m$^2$ | 32.8237 |
| **Measurement error correlation width (in 15 m pixels)** | | | | |
| 1 | -//- | $\sigma_{Corr01}^2 = 0.1525$ | 0.00227 | 67.0331 |

In total, 3337 $\hat{\sigma}_e^2$ estimates and 452 $\hat{\sigma}_{Corr}^2$ estimates were obtained by the mvcNI+fBm. They cover predictor variation intervals from 4 to 7 for $N_{stk}$ and from 1000 to about 3800 m for $Z$. No outliers were detected for both error variance and correlation width.

TABLE III
THE RESULTS OF ELEVATION MEASUREMENT ERROR PARAMETER ESTIMATION OBTAINED WITH THE MVCNI+FBM FOR THE ALOS 3D WORLD, 30 M

| Model type | $R^2$ | Parameters | Parameter SDs | t-stats |
|---|---|---|---|---|
| **Measurement error variance** | | | | |
| $1+N_{stk}^{-1}$ | 0.03179 | $\sigma_{Var01}^2 = 0.3340$ m$^2$ | 0.00407 m$^2$ | 82.0500 |
| | | $\sigma_{Var02}^2 = 0.2770$ m$^2$ | 0.02404 m$^2$ | 11.5208 |
| **Measurement error correlation width (in 90 m pixels)** | | | | |
| 1 | -//- | $\sigma_{Corr01}^2 = 0.2556$ | 0.00139 | 183.6447 |

In total, 8126 $\hat{\sigma}_e^2$ estimates and 5645 $\hat{\sigma}_{Corr}^2$ estimates were obtained by the mvcNI+fBm. They cover predictor variation intervals from 3 to 14 for $N_{stk}$ and from -400 to about 5000 m for $Z$. The percentage of detected outliers is about 0.3% for the error variance and about 1.8% for error correlation width.

been applied to build bivariate models of ASTER GDEM2 and ALOS World 3D (both with 5 m and 30 m spatial resolution) elevation measurement error. The two predictors in these models are number of stereo pairs (responsible for stacking procedure influence) and elevation itself (responsible to epipolar line error influence). These models have been found consistent with the accuracy analysis results published in the available literature for GDEM2 and ALOS World 3D data. Not previously reported in the literature, our analysis reveals the epipolar line error as a very important factor responsible for the GDEM2 quality degradation for high elevations. The derived regression models for GDEM2, AW3D30, and AW3D can be used to predict elevation measurement error parameters for low relief areas (with elevation SD from 0 to about 15 m).

While the proposed no-reference approach to DEM accuracy analysis cannot provide information on DEM bias, the detailed study of DEM measurement error SD and correlation width can see variety of applications including analysis of LIDAR or RADAR-derived raster DEMs, direct efficiency comparison of different approaches for stereo matching or interferometric phase reconstruction, analysis of sensor-related (e.g. jitter causing the epipolar line error of GDEM2) components in elevation measurement error. Future work is seen in the direction of using additional observable predictors (especially related to vegetation cover). Another direction is to overcome inability of characterizing DEM error in high relief areas.